\documentclass[conference]{IEEEtran}
\IEEEoverridecommandlockouts
\usepackage{amsmath,amssymb,amsfonts}
\usepackage{algorithmic}
\usepackage{graphicx}
\usepackage{textcomp}
\usepackage{xcolor}
\usepackage{multirow} 
\usepackage{subcaption}
\usepackage[normalem]{ulem}
\usepackage[a4paper, total={184mm,239mm}]{geometry}
\def\BibTeX{{\rm B\kern-.05em{\sc i\kern-.025em b}\kern-.08em
    T\kern-.1667em\lower.7ex\hbox{E}\kern-.125emX}}
\usepackage{biblatex}
\begin{document}

\title{A Unified Hyperparameter Optimization Pipeline for Transformer-Based Time Series Forecasting Models
\thanks{This work was funded by the Luxembourg National Research Fund (Fonds National de la Recherche - FNR), Grant ID 15748747 and Grant ID C21/IS/16221483/CBD. We thank to Jean-Francois Nies from Luxembourg Institute of Science and Technology (LIST) for proofreading. We also thank our TPU Fernand for its hard work.
}
}

\makeatletter
\newcommand{\linebreakand}{%
  \end{@IEEEauthorhalign}
  \hfill\mbox{}\par
  \mbox{}\hfill\begin{@IEEEauthorhalign}
}
\makeatother
\author{
\IEEEauthorblockN{1\textsuperscript{st} Jingjing Xu}
\IEEEauthorblockA{
\textit{FSTM/DCS} \\
\textit{University of Luxembourg}\\
Esch-sur-Alzette, Luxembourg \\
\url{https://orcid.org/0000-0002-0012-3911}
}
\and
\IEEEauthorblockN{2\textsuperscript{rd} Caesar Wu}
\IEEEauthorblockA{
\textit{SnT} \\
\textit{University of Luxembourg}\\
Esch-sur-Alzette, Luxembourg  \\
\url{https://orcid.org/0000-0002-2792-6466}
}
\and
\IEEEauthorblockN{3\textsuperscript{th} Yuan-Fang Li}
\IEEEauthorblockA{\textit{Faculty of Information Technology} \\
\textit{Monash University}\\
Victoria, Australia \\
\url{https://orcid.org/0000-0003-4651-2821}}
\linebreakand
\IEEEauthorblockN{4\textsuperscript{th} Grégoire Danoy}
\IEEEauthorblockA{\textit{FSTM/DCS \& SnT} \\
\textit{University of Luxembourg}\\
Esch-sur-Alzette, Luxembourg \\
\url{https://orcid.org/0000-0001-9419-4210}}
\and
\IEEEauthorblockN{5\textsuperscript{th} Pascal Bouvry}
\IEEEauthorblockA{\textit{FSTM/DCS \& SnT} \\
\textit{University of Luxembourg}\\
Esch-sur-Alzette, Luxembourg \\
\url{https://orcid.org/0000-0001-9338-2834}}
}

\maketitle

\begin{abstract}

Transformer-based models for time series forecasting (TSF) have attracted significant attention in recent years due to their effectiveness and versatility. However, these models often require extensive hyperparameter optimization (HPO) to achieve the best possible performance, and a unified pipeline for HPO in transformer-based TSF remains lacking. In this paper, we present one such pipeline and conduct extensive experiments on several state-of-the-art (SOTA) transformer-based TSF models. These experiments are conducted on standard benchmark datasets to evaluate and compare the performance of different models, generating practical insights and examples. Our pipeline is generalizable beyond transformer-based architectures and can be applied to other SOTA models, such as Mamba and TimeMixer, as demonstrated in our experiments. The goal of this work is to provide valuable guidance to both industry practitioners and academic researchers in efficiently identifying optimal hyperparameters suited to their specific domain applications. The code and complete experimental results are available on GitHub\footnote{\url{https://github.com/jingjing-unilu/HPO_transformer_time_series}}.
\end{abstract}

\section{Introduction}
\label{intro}
Time series forecasting (TSF) is important for decision making across diverse practical domains, making it a continuously evolving field. Over time, TSF models have progressed from classic approaches, such as auto-regressive-moving-average (ARMA) models and exponential smoothing, to more sophisticated deep learning models, largely due to rapid advancements in computational capabilities~\cite{de200625,petropoulos2022forecasting}. 
Among these advancements, deep learning models, particularly transformer models, have demonstrated significant potential in improving the accuracy and efficiency of TSF. However, these models often depend on a wide range of hyperparameters, and optimizing them typically requires substantial expertise. Thereby, these models increase the technical barriers for users attempting to apply these models to different datasets with varied hyperparameter configurations~\cite{feurer2019hyperparameter,yu2020hyper,meisenbacher2022review}. As a result, hyperparameter optimization (HPO) become increasingly critical for ensuring the effective use of transformer models in TSF.

In the context of machine learning and deep learning, HPO refers to the process of selecting an optimal set of hyperparameters for a model to minimize a predefined loss function with given specific dataset~\cite{claesen2015hyperparameter}.While several studies have explored HPO for various machine learning models and TSF applications, research specifically focused on HPO for transformer-based TSF models remains limited. To address this gap, we introduce a unified HPO pipeline designed specifically for transformer-based TSF models. Additionally, we evaluate several state-of-the-art (SOTA) models on standard datasets to provide practical insights and examples of the pipeline’s effectiveness.

The structure of the paper is as follows: Sec.~\ref{2_back} reviews the background and related work; Sec.~\ref{3_exp} details the experimental setup; Sec.~\ref{4_resul} presents the results and analysis; and Sec.~\ref{5_conclusion} concludes the paper with future directions. Our aim is to enhance the reproducibility, fairness, and performance of TSF models across diverse datasets and domains. The key contributions of this work are as follows:
\begin{enumerate}
    \item We introduce a Hyper-Parameter Tuning pipeline specifically designed for transformer-based and other TSF models.
	\item We benchmark standard datasets using various SOTA TSF models, including transformer-based models, Mamba~\cite{gu2023mamba}, and TimeMixer~\cite{wang2023timemixer}.
	\item We perform a comprehensive analysis of the experimental results.
\end{enumerate}

\section{Background and Related Work}
\label{2_back}


\subsection{Transformer-based Time Series Forecasting}

The Transformer model has emerged as a good option for time series forecasting (TSF) due to its excellent ability to capture long-range dependencies. Several transformer-based forecasting models have been developed to address various forecasting challenges. Informer~\cite{zhou2021informer} and Autoformer~\cite{wu2021autoformer} pioneers the adaptation of transformer components for time series applications. Subsequently, the Crossformer model~\cite{zhang2022crossformer} focuses on capturing cross-time and cross-dimensional dependencies to enhance multivariate time series forecasting. Meanwhile, the Non-stationary Transformer~\cite{liu2022non} addresses issues related to stationarity in forecasting tasks. Following this, PatchTST~\cite{nietime} employs patching and channel-independent architectures to effectively capture both local and longer lookback information. The iTransformer~\cite{liuitransformer} reconfigures the traditional transformer structure, offering an alternative approach for time series forecasting. Comprehensive surveys~\cite{ahmed2023transformers,wen2023transformers,xu2024survey} provide an overview of transformer models tailored for TSF. Additionally, researchers have begun to apply transformer-based TSF methods in finance~\cite{wu2024trustworthy,wu2024strategic}. Furthermore, emerging techniques and models for time series forecasting include graph representation learning~\cite{jin2022multivariate}, Large Language Models (LLMs)~\cite{jintime}, Mamba~\cite{gu2023mamba}, and TimeMixer~\cite{wang2023timemixer}.

\subsection{HPO for Transformer}
Hyperparameter optimization (HPO) search algorithms include Grid Search, Random Search, Bayesian Optimization (BO), Tree Parzen Estimators (TPE), and others~\cite{yu2020hyper,talbi:hal-02570804}. HPO is a critical component of machine learning, as it enables models to select the optimal set of hyperparameters to maximize performance on a given dataset. Some researchers have addressed HPO in the context of common machine learning models~\cite{claesen2015hyperparameter,yang2020hyperparameter}, and discussions surrounding HPO for deep learning and neural networks have also emerged. Generally, hyperparameters in deep learning can be classified into two categories: those related to model architecture and those associated with training and optimization. In works related to model architecture, a recent survey~\cite{chitty2023neural} explores neural architecture search (NAS) benchmarks, highlighting the need for efficient search algorithms. Additionally, another survey~\cite{chitty2022neural} summarizes the NAS landscape for Transformers and their associated architectures, specifically discussing HPO in autotransformer~\cite{ren2022autotransformer} for time series classification tasks. However, the exploration of HPO for Transformers in time series forecasting tasks remains insufficient.

\subsection{HPO for Time Series Forecasting}
HPO is crucial for improving forecast performance and mitigating overfitting issues in time series forecasting. A review~\cite{meisenbacher2022review} identifies hyperparameter optimization (HPO) as one of the five key components of the time series forecasting pipeline, concluding that the grid search method is the most widely used in automated forecasting frameworks. Additionally, evolutionary optimization and Bayesian Optimization are often employed in high-complexity training processes. The paper~\cite{dhake2023algorithms} presents hyperparameter tuning algorithms specifically for Long Short-Term Memory (LSTM) networks, aiming to efficiently determine the optimal set of hyperparameters. Furthermore, another paper~\cite{singh2024distributed} proposes a distributed HPO approach for time series forecasting based on electricity dataset. Researchers~\cite{hanifi2024advanced} investigate the three hyperparameter tuning toolkits Scikit-opt, Optuna and Hyperopt, and then apply these toolkits to Convolutional Neural Networks (CNN) and LSTM models for wind power prediction. Recently, the paper~\cite{stefenon2024hypertuned} introduced an automatic hyperparameter tuning framework for the Temporal Fusion Transformer (AutoTFT) model (for multi-horizon time series forecasting). Despite these advancements, research focused on HPO for various transformer-based time series forecasting models across different model on datasets remains limited.

\section{Experiments}
\label{3_exp}


In this study, we perform hyperparameter optimization for long-term time series forecasting. The primary objective is to identify a set of hyperparameter values that minimizes forecasting errors across different models.
\subsection{Dataset and Metrics}
We utilize widely-used open-source datasets for long-term time series forecasting, including ETTh1, Weather and Electricity. The evaluation metrics employed in this experiment are Mean Squared Error (MSE) and Mean Absolute Error (MAE), which are commonly used to assess model performance, as noted in the survey~\cite{xu2024survey}. A summary of the datasets is provided in Tab.~\ref{dataset}, and detailed descriptions of the datasets can be found in related works~\cite{wu2021autoformer,zhou2021informer}.

\begin{table}[!ht]
    \centering
    \caption{Summary of three datasets.}
    \label{dataset}
    \begin{tabular}{c|c|c|c}
     \hline
        Datasets & ETTh1 & Weather & Electricity (ECL)  \\ 
        \hline
        Variables & 7 & 21 & 321  \\ 
        \hline
        Timesteps & 17420 & 52696 & 26304  \\
    \hline
    \end{tabular}
\end{table}

\begin{table*}[!h]
\caption{Parameters for Different Models. NOTE: 1)DF:Default}
\label{model_parameters}
\centering
\resizebox{\linewidth}{!}{
\begin{tabular}{l | l | c | cccc | cccc | cccc | cccc | c | c}
\hline
\multicolumn{1}{c|}{\multirow{2}{*}{Parameter}} & \multicolumn{1}{c|}{\multirow{2}{*}{Module}} & \multicolumn{1}{c|}{\multirow{2}{*}{Default}} & \multicolumn{4}{c|}{PatchTST}            & \multicolumn{4}{c|}{Crossformer}         & \multicolumn{4}{c|}{Autoformer}          & \multicolumn{4}{c|}{Non-stationary\_Transformer} & Lowest & Highest \\
\multicolumn{1}{c|}{}                           & \multicolumn{1}{c|}{}                        & \multicolumn{1}{c|}{}                         & ETTh1 & weather & electricity & traffic & ETTh1 & weather & electricity & traffic & ETTh1 & weather & electricity & traffic & ETTh1   & weather   & electricity   & traffic  & Value  & Value   \\
\hline
data                                           & \# data loader                              & ETTm1                                       & ETTh1 & custom  & custom      & custom  & ETTh1 & custom  & custom      & custom  & ETTh1 & custom  & custom      & custom  & ETTh1   & custom    & custom        & custom   & -      & -       \\
features                                       & \# data loader                              & 'M'                                          & M     & M       & M           & M       & M     & M       & M           & M       & M     & M       & M           & M       & M       & M         & S             & M        & -      & -       \\
seq\_len                                       & \# forecasting task                         & 96                                           & 96    & 96      & 96          & 96      & 96    & 96      & 96          & 96      & 96    & 96      & 96          & 96      & 96      & 96        & 96            & 96       & 96     & 96      \\
label\_len                                     & \# forecasting task                         & 48                                           & 48    & 48      & 48          & 48      & 48    & 48      & 48          & 96      & 48    & 48      & 48          & 48      & 48      & 48        & 48            & 48       & 48     & 96      \\
pred\_len                                      & \# forecasting task                         & 96                                           & 96    & 96      & 96          & 96      & 96    & 96      & 96          & 96      & 96    & 96      & 96          & 96      & 96      & 96        & 96            & 96       & 96     & 96      \\
e\_layers                                      & \# model define                             & 2                                            & 1     & 2       & 2           & 2       & 2     & 2       & 2           & 2       & 2     & 2       & 2           & 2       & 2       & 2         & 2             & 2        & 1      & 2       \\
d\_layers                                      & \# model define                             & 1                                            & 1     & 1       & 1           & 1       & 1     & 1       & 1           & 1       & 1     & 1       & 1           & 1       & 1       & 1         & 1             & 1        & 1      & 1       \\
factor                                         & \# model define                             & 1                                            & 3     & 3       & 3           & 3       & 3     & 3       & 3           & 3       & 3     & 3       & 3           & 3       & 3       & 3         & 3             & 3        & 1      & 3       \\
enc\_in                                        & \# model define                             & 7                                            & 7     & 21      & 321         & 862     & 7     & 21      & 321         & 862     & 7     & 21      & 321         & 862     & 7       & 21        & 321           & 862      & 7      & 862     \\
dec\_in                                        & \# model define                             & 7                                            & 7     & 21      & 321         & 862     & 7     & 21      & 321         & 862     & 7     & 21      & 321         & 862     & 7       & 21        & 321           & 862      & 7      & 862     \\
c\_out                                         & \# model define                             & 7                                            & 7     & 21      & 321         & 862     & 7     & 21      & 321         & 862     & 7     & 21      & 321         & 862     & 7       & 21        & 321           & 862      & 7      & 862     \\
d\_model                                       & \# model define                             & 512                                          & DF    & DF      & DF          & 512     & DF    & 32      & 256         & DF      & DF    & DF      & DF          & DF      & 128     & DF        & 2048          & DF       & 32     & 2048    \\
d\_ff                                          & \# model define                             & 2048                                         & DF    & DF      & DF          & 512     & DF    & 32      & 512         & DF      & DF    & DF      & DF          & DF      & DF      & DF        & DF            & DF       & 32     & 2048    \\
top\_k                                         & \# model define                             & 5                                            & DF    & DF      & DF          & 5       & DF    & 5       & 5           & 5       & DF    & DF      & DF          & DF      & DF      & DF        & DF            & DF       & 5      & 5       \\
n\_heads                                       & \# model define                             & 8                                            & 2     & 4       & DF          & DF      & DF    & DF      & DF          & 2       & DF    & DF      & DF          & DF      & DF      & DF        & DF            & DF       & 2      & 8       \\
dropout                                        & \# model define                             & 0.1                                          & DF    & DF      & DF          & DF      & DF    & DF      & DF          & DF      & DF    & DF      & DF          & DF      & DF      & DF        & DF            & DF       & 0.1    & 0.1     \\
des                                            & \# optimization                             & 'test'                                       & Exp   & Exp     & Exp         & Exp     & Exp   & Exp     & Exp         & Exp     & Exp   & Exp     & Exp         & Exp     & Exp     & Exp       & Exp           & Exp      &  -      &  -       \\
itr                                            & \# optimization                             & 1                                            & DF     & DF       & DF           & DF       & DF     & DF       & DF           & DF       & DF     & DF       & DF           & DF       & DF       & DF         & DF             & DF        & 1      & 1       \\
train\_epochs                                  & \# optimization                             & 10                                           & DF    & 3       & DF          & DF      & DF    & 1       & DF          & DF      & DF    & 2       & DF          & 3       & DF      & 3         & DF            & 3        & 1      & 10      \\
batch\_size                                    & \# optimization                             & 32                                           & DF    & 128     & 16          & 4       & DF    & DF      & 16          & 4       & DF    & DF      & DF          & DF      & DF      & DF        & DF            & DF       & 4      & 128     \\
learning\_rate                                 & \# optimization                             & 0.0001                                           & DF    & DF     & DF          & DF       & DF    & DF      & DF          & DF       & DF    & DF      & DF          & DF      & DF      & DF        & DF            & DF       & 0.0001       & 0.0001      \\
p\_hidden\_dims                                & \# de-stationary projector params           & {[}128,128{]}                                & DF    & DF      & DF          & DF      & DF    & DF      & DF          & DF      & DF    & DF      & DF          & DF      & 256     & 256       & 256           & 128      & 128    & 256     \\
p\_hidden\_layers                              & \# de-stationary projector params           & 2                                            & DF    & DF      & DF          & DF      & DF    & DF      & DF          & DF      & DF    & DF      & DF          & DF      & 2       & 2         & 2             & 2        & 2      & 2      \\
\hline
\end{tabular}
}
\end{table*}

\subsection{Environment and Configuration}
All experiments in this study were conducted on a single Nvidia TU02 GPU. 
\subsubsection{Model and its Setting}
 For our case study, we randomly selected four transformer-based TSF models: Autoformer~\cite{wu2021autoformer}, Crossformer~\cite{zhang2022crossformer}, Non-Stationary Transformer~\cite{liu2022non}, and PatchTST~\cite{nietime}. Additionally, we include other state-of-the-art (SOTA) models, such as Mamba~\cite{gu2023mamba} and TimeMixer~\cite{wang2023timemixer}, for comparison. For model settings, we selected the long-term forecasting task as the primary focus of this paper. Additionally, while we arbitrarily chose a prediction length of 96 as a use case, other prediction lengths (192, 336, 720) can be implemented in a similar fashion. In addition, the evaluation metrics used are Mean Squared Error (MSE) and Mean Absolute Error (MAE), which are standard measures for model performance. 

\subsubsection{HPO Setting}
During the hyperparameter tuning process, the model with the lowest validation loss (MSE) is defined as the best-performing model. Each model undergo 20 trials on each dataset to tune hyperparameters. We utilize OptunaSearch (a variant of Tree-structured Parzen Estimator, TPE) as an example of search algorithms. Our pipeline also supports commonly used search algorithms such as random search, grid search, and Bayesian optimization. As a hyperparameter tuning tool, we utilized Ray Tune\footnote{https://github.com/ray-project/ray}, a scalable tool designed to optimize model performance efficiently. Additionally, Weights \& Biases\footnote{https://github.com/wandb/wandb} is applied to visualize the hyperparameter tuning processes and outcomes. The code of this paper is built upon the Time Series Library~\cite{wu2023timesnet}.

\begin{figure}[!ht]
    \centering
    \includegraphics[width=0.95\linewidth]{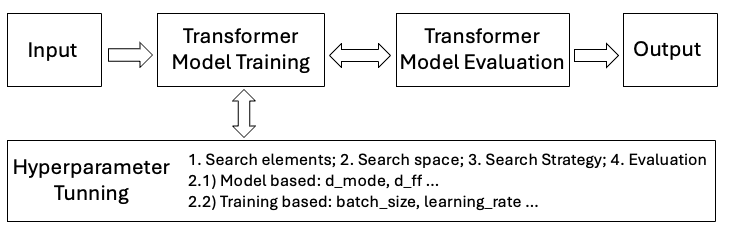}
    \caption{HPO pipeline for Transformer-based forecasting}
    \label{fig:HPO_pipeline}
\end{figure}
\subsection{HPO Pipeline}

The hyperparameter optimization (HPO) pipeline is depicted in Fig.~\ref{fig:HPO_pipeline}. The process begins with feeding data into the model, followed by the simultaneous execution of model training and hyperparameter tuning. Once the training is completed, the model is evaluated, and the results are output. In line with the Neural Architecture Search (NAS) framework, four key components are integral to the hyperparameter tuning process: collecting primitive search elements (hyperparameters), designing the search space, selecting the search algorithm, and evaluating performance to determine the optimal model or network~\cite{chitty2022neural}. OptunaSearch is employed in the experiments as the search algorithm, and the evaluation of results is discussed in Sec.~\ref{4_resul}. Therefore, we emphasize the collection of primitive hyperparameters and the design of the search space in this section. 

\begin{table}[!ht]
\caption{Common Parameters in this Experiments}
\label{exp_parameters}
\centering
\resizebox{\linewidth}{!}{
\begin{tabular}{l | c | c | c | c| c}
\hline
\multirow{2}{*}{Parameter} & \multirow{2}{*}{Module}          & \multirow{2}{*}{Default} & Lowest & Highest & \multirow{2}{*}{Searching Space}                              \\
      &            &   & Value & Value &                           \\
\hline
d\_ff          & model define & 2048    & 32     & 2048    & {[}16,32,64,128,256,512,1024,2048,4096{]} \\
d\_layers      & model define & 1       & 1      & 1       & {[}1,2{]}                               \\
d\_model       & model define & 512     & 32     & 2048    & {[}16,32,64,128,256,512,1024,2048,4096{]} \\
e\_layers      & model define & 2       & 1      & 2       & {[}1,2,3{]}                               \\
factor         & model define & 1       & 1      & 3       & {[}1,2,3,4{]}                             \\
n\_heads       & model define & 8       & 2      & 8       & {[}2,4,8,16{]}                            \\
batch\_size    & optimization & 32      & 4      & 128     & {[}4, 16, 32, 64, 128, 256{]}             \\
learning\_rate & optimization & 0.0001  & -      & -       & {[}0.00001, 0.0001, 0.001{]}              \\
train\_epochs  & optimization & 10      & 1      & 10      & {[}1,2,3,4,5,6,7,8,9,10,11{]}     \\
\hline
\end{tabular}
}
\end{table}

\subsubsection{Hyperparameters and Search Spaces} 
We first gather all hyperparameters from the Time Series Library\footnote{https://github.com/thuml/Time-Series-Library}, and then review the parameters used across different models as shown in Tab.~\ref{model_parameters} (Mamba and TimeMixer follow the same lowest and highest value). Most of the parameters are assigned default values. For the hyperparameter tuning experiments, we select common parameters shared by the models. The selected parameters and corresponding search spaces are presented in Tab.~\ref{exp_parameters}. These hyperparameters are divided into two categories: ``model define'' and ``optimization (training)'' groups. The `optimization (training)'' group includes parameters that influence the speed of convergence, while the ``model define'' group comprises hyperparameters that determine the learning capacity of the model. The parameters in the ``model define'' group are illustrated in Fig.~\ref{fig:parameters_in_transformer}. To define the search space, we first identified the minimum and maximum values for each parameter across models. The search space range is then extended by setting the lower bound one step below the minimum value and the upper bound one step above the maximum value.
\begin{figure*}[!ht]
    \centering
    \includegraphics[width=0.95\linewidth]{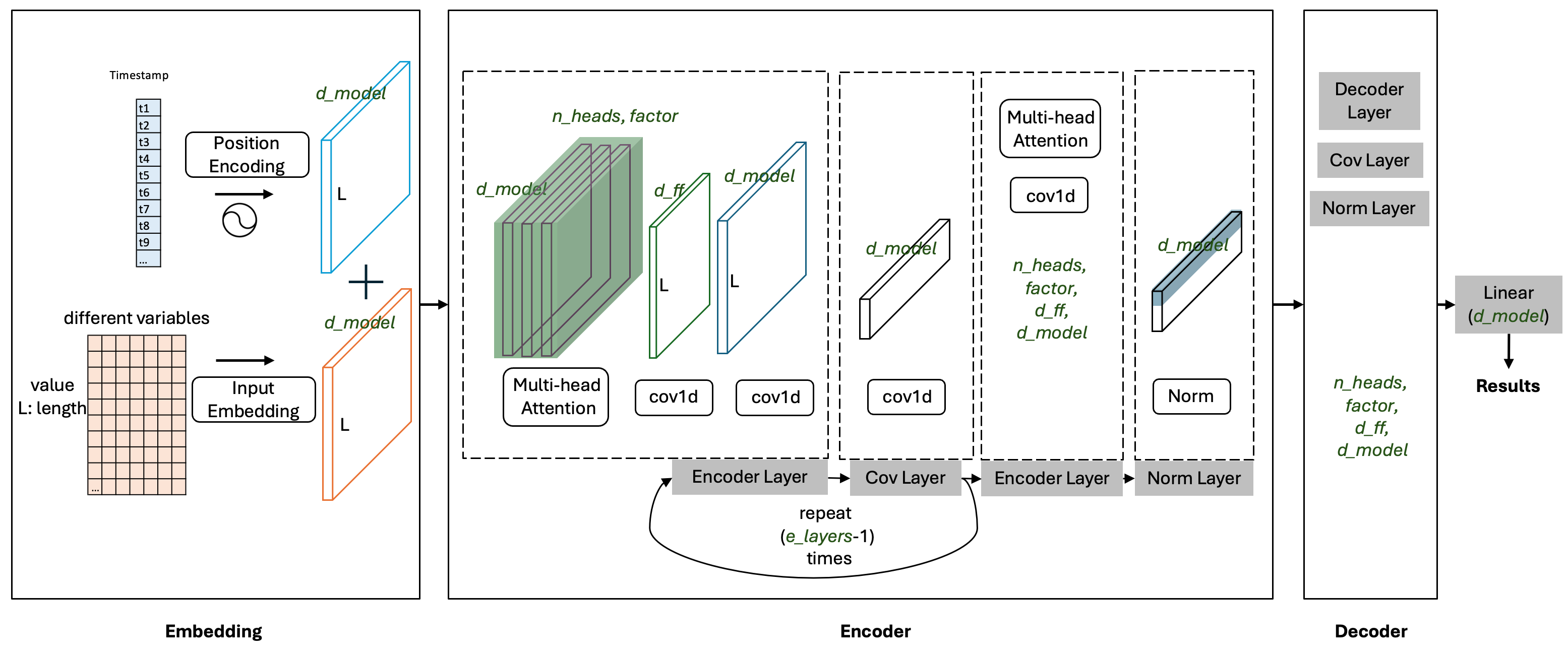}
    \caption{Common Parameters in the Transformer}
    \label{fig:parameters_in_transformer}
\end{figure*}

\section{Results and Analysis}
\label{4_resul}
In our experiments, each model is subjected to 20 trials per dataset, employing the OptunaSearch algorithm for hyperparameter optimization. A total of 357 trials are conducted (6 models * 20 trials * 3 datasets, minus 3 unexecuted cases). We analyze these experiments by examining the best performance of each model and investigating the behavior of hyperparameters. 

\subsection{Best results on different datasets}

Tab.~\ref{best_results} presents the best results for each model across the datasets. \textbf{bold} numbers indicate the best performance, while \underline{underline} numbers denote the second-best performance. For the ETTh1 dataset, TimeMixer achieved the best results. In contrast, for the Weather and ECL datasets, Crossformer demonstrates better performance even it is a time-consuming model. The learning curve in Fig.~\ref{fig:best_perfomance_loss} illustrates the performance behavior of each model. The details are as follows.

\begin{table}[!ht]
\caption{Best result of different models on different datasets}
\label{best_results}
\centering
\resizebox{\linewidth}{!}{
\begin{tabular}{c | c c c | c c c | c c c}
\hline
{\multirow{2}{*}{Model}} & \multicolumn{3}{c|}{ETTh1}  & \multicolumn{3}{c|}{Weather} & \multicolumn{3}{c}{Electricity (ECL)}  \\
\cline{2-10}
                                            & MSE              & MAE            & Time      & MSE               & MAE        & Time         & MSE                & MAE    & Time\\
\hline
PatchTST                                   & \underline{0.3852}     & 0.3974    & 654       & 0.1735            & 0.2158     & 621         & 0.1739          & 0.2699        & 532 \\
Crossformer                                & 0.4149           & 0.4412          & 62        & \textbf{0.1527}   & 0.2278     & 3458        & \textbf{0.1347} & 0.2319       & 8647\\
Autoformer                                 & 0.4466           & 0.4559          & 136       & 0.2857            & 0.3552     & 190         & 0.1981          & 0.3121        & 3330\\
Nons. Trans.                               & 0.5515           & 0.5126          & 106       & 0.1867            & 0.2342     & 225         & \underline{0.1683}  & 0.2729    & 284 \\
\hline
Mamba                                      & 0.4701           & 0.4416          & 499       & 0.1942            & 0.2413     & 481         & 0.1691          & 0.2719        & 2643\\
TimeMixer                                  & \textbf{0.3815}  & 0.3967          & 128       & \underline{0.1614}   & 0.2088  & 4626         & 0.1747          & 0.2648       & 2994\\
\hline
\end{tabular}
}
\end{table}

\begin{figure*}[!h]
    \centering
    \begin{subfigure}[b]{0.32\linewidth}
         \includegraphics[clip, trim=0.5cm 20cm 6.5cm 1.5cm, width=1.00\textwidth]{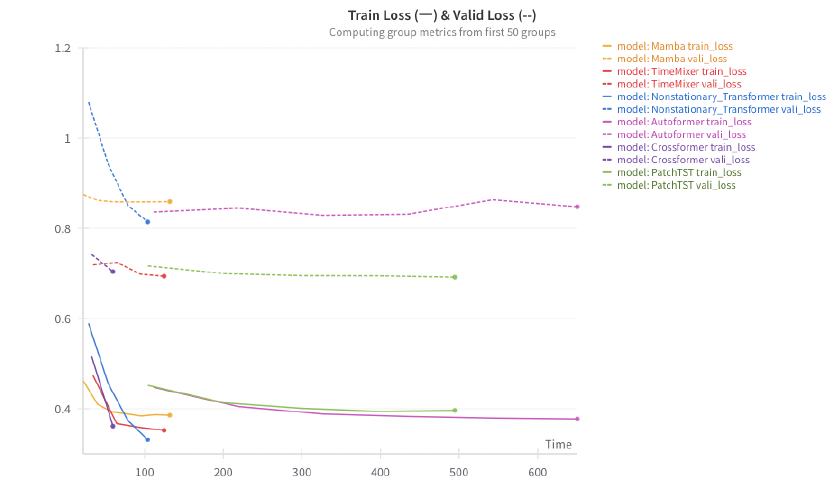}
        \caption{ETTh1}
        \label{fig:ETTh1_best}
    \end{subfigure}
    \hfill
    \begin{subfigure}[b]{0.32\linewidth}
         \includegraphics[clip, trim=0.5cm 11cm 6.5cm 10.5cm, width=1.00\textwidth]{conference-template-A4-latex/train_vali_loss.pdf}
        \caption{Weather}
        \label{fig:Weather_best}
    \end{subfigure}
    \hfill
    \begin{subfigure}[b]{0.32\linewidth}
        \includegraphics[clip, trim=0.5cm 2cm 6.5cm 19cm, width=1.00\textwidth]{conference-template-A4-latex/train_vali_loss.pdf}
        \caption{ECL}
        \label{fig:ECL_best}
    \end{subfigure}
\caption{Training loss and validation loss on each model's best performance case in experiments}
\label{fig:best_perfomance_loss}
\end{figure*}

\subsubsection{ETTh1 dataset}

Fig.~\ref{fig:ETTh1_best} shows that in all models, including the best-performing TimeMixer, the validation loss exceeds the training loss, suggesting underfitting. This indicates that the models have not sufficiently captured the underlying patterns in the training data. From a hyperparameter tuning perspective, increasing the number of epochs or the model’s complexity could improve performance. From a data perspective, noise reduction and feature engineering may also help~\cite{xu2023transformer, GeeksforGeeks_2024, xu2024survey, wen2021time}.

\subsubsection{Weather dataset}
Crossformer achieved the best performance on the Weather dataset, but it was trained for only one epoch, suggesting the need for further hyperparameter tuning. Fig.~\ref{fig:Weather_best} indicates that most models are underfitting and require additional tuning. However, Autoformer showed slight overfitting, as its validation loss began to increase after initially decreasing. Overfitting suggests that the model cannot generalize well to new data, so reducing model complexity, employing early stopping, or using dropout techniques would be beneficial in future hyperparameter tuning efforts.

\subsubsection{ECL dataset}
On the ECL dataset, Crossformer again demonstrated the best performance (see Table \ref{best_results}). As shown in Fig.~\ref{fig:ECL_best}, the model performed well, with a small difference between training and validation loss after both decreased. While the results are promising, further hyperparameter tuning can still be conducted to improve the model capability and performance.

\subsection{Hyperparmater and Model Metric}

To understand the relationship between hyperparameters and model performance, we map hyperparameter values to validation loss using parallel coordinate plots (lower validation loss means better performance). This approach help us identify the best and worst hyperparameter ranges for each model in further tuning. However, we notice that some models crash due to out-of-memory (OOM) errors, as shown in Tab.~\ref{Tab:OOM_rate}, particularly with the ECL dataset (which has 321 variables). The high number of variables make models such as Crossformer, PatchTST, and TimeMixer sensitive to hyperparameters, resulting in OOM errors. Mamba also encountered a high failure rate, with nearly 50\% of trials crashing. To mitigate these OOM errors in the future, we analyze the OOM case in ECL firstly and then continue the analysis of ETTh1 and Weather dataset. All trials' results and parallel coordinate plots are available via GitHub.

\begin{table}[h!]
\centering
\caption{Out-of-Memory Rate}
\label{Tab:OOM_rate}
\begin{tabular}{c|c|c|c}
\hline
Models                                         & ETTh1 & Weather & ECL  \\
\hline
Autoformer                                     & 10\%  & 10\%    & 10\% \\
Crossformer                                    & 5\%   & 25\%    & 35\% \\
Nons.\_Trans. & 10\%  & 0\%     & 10\% \\
PatchTST                                       & 0\%   & 0\%     & 40\% \\
\hline
Mamba                                          & 50\%  & 53\%    & 40\% \\
TimeMixer                                      & 10\%  & 40\%    & 85\% \\
\hline
\end{tabular}
\end{table}

\subsubsection{Dataset: Electricity (ECL)}
During hyperparameter tuning, all models experience crash cases on the ECL dataset. Tab.~\ref{Tab:OOM_ECL} summarizes the models that crashed under the training and model-related parameters ($batch\_size$, $d\_model$, $d\_ff$). For instance, the Autoformer model encountered OOM errors when the $batch\_size$ is set to 256 while $max(d\_model, d\_ff) \geq 2048$ and $min(d\_model, d\_ff) \geq 16$ (both $d\_model$ and $d\_ff$ influence the model size). Therefore, hyperparameter tuning process requires more GPU memory or smaller batch sizes~\cite{popel2018training} for the dataset like ECL. 

\begin{table}[h!]
\centering
\caption{Out-of-Memory Case (ECL)}
\label{Tab:OOM_ECL}
\resizebox{0.9\linewidth}{!}{
\begin{tabular}{c|c|c|c}
\hline
Model         & Batch Size         & Max(d\_model, d\_ff) & Min(d\_model, d\_ff) \\
\hline
Autoformer    & 256 & \textgreater{}=2048  & \textgreater{}=16    \\
\hline
Crossformer   & 128 & \textgreater{}=2048  & \textgreater{}=256   \\
Crossformer   & 256 & \textgreater{}=2048  & \textgreater{}=32    \\
\hline
Nons.\_Trans. & 128 & \textgreater{}=4096  & \textgreater{}=16    \\
\hline
PatchTST      & 32  & \textgreater{}=4096  & \textgreater{}=1024  \\
PatchTST      & 64  & \textgreater{}=4096  & \textgreater{}=512   \\
PatchTST      & 128 & \textgreater{}=2048  & \textgreater{}=64    \\
PatchTST      & 256 & \textgreater{}=1024  & \textgreater{}=32    \\
\hline
Mamba         & 4   & \textgreater{}=2048  & \textgreater{}=64    \\
Mamba         & 16  & \textgreater{}=2048  & \textgreater{}=32    \\
Mamba         & 32  & \textgreater{}=4096  & \textgreater{}=32    \\
Mamba         & 64  & \textgreater{}=1024  & \textgreater{}=32    \\
Mamba         & 256 & \textgreater{}=4096  & \textgreater{}=16    \\
\hline
TimeMixer     & 16  & \textgreater{}=512   & \textgreater{}=128   \\
TimeMixer     & 32  & \textgreater{}=256   & \textgreater{}=32    \\
TimeMixer     & 64  & \textgreater{}=128   & \textgreater{}=64    \\
TimeMixer     & 128 & \textgreater{}=256   & \textgreater{}=32    \\
TimeMixer     & 256 & \textgreater{}=512   & \textgreater{}=128   \\
\hline
\end{tabular}
}
\end{table}

\subsubsection{Dataset: ETTh1}

\begin{figure}[]
    \centering
    \includegraphics[width=0.95\linewidth]{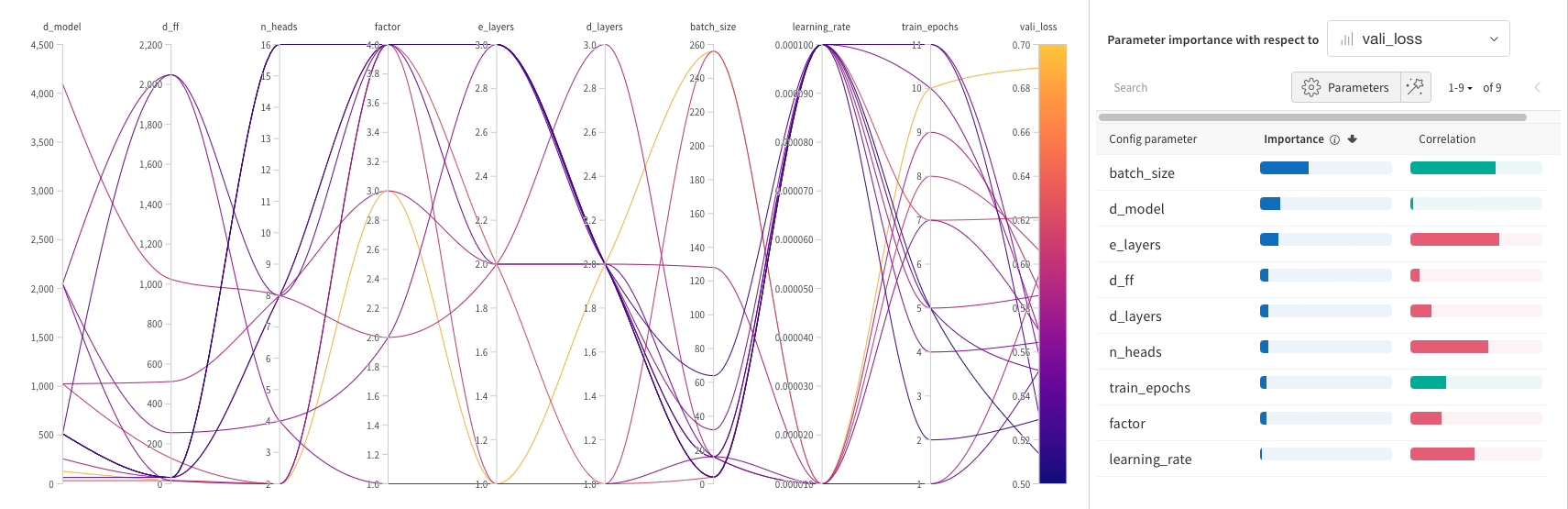}
    \caption{Parallel coordinates plot on Weather dataset: Autoformer without outlier}
    \label{fig:autoformer_zoomin_weather_parallel_plot}
\end{figure}

\begin{figure}[h!]
\centering
  \begin{subfigure}[b]{0.95\linewidth}
    \includegraphics[width=\textwidth]{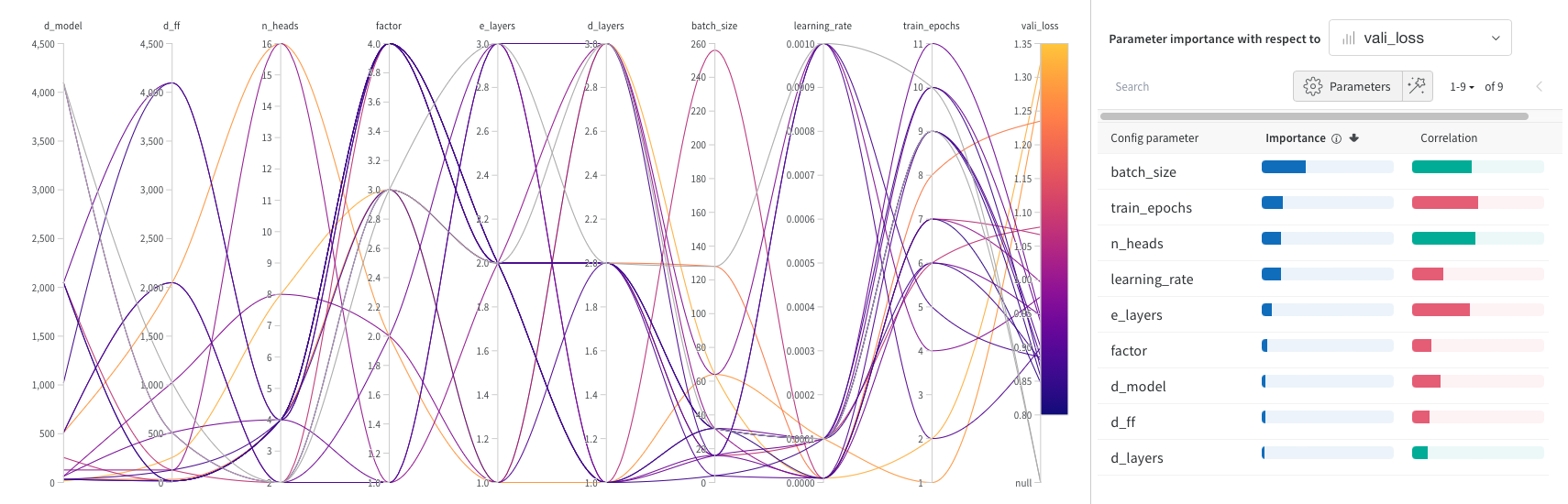}
    \caption{Autoformer}
    \label{fig:autoformer_parallel_coordinates_plot}
  \end{subfigure}
\hfill
  \begin{subfigure}[b]{0.95\linewidth}
    \includegraphics[width=\textwidth]{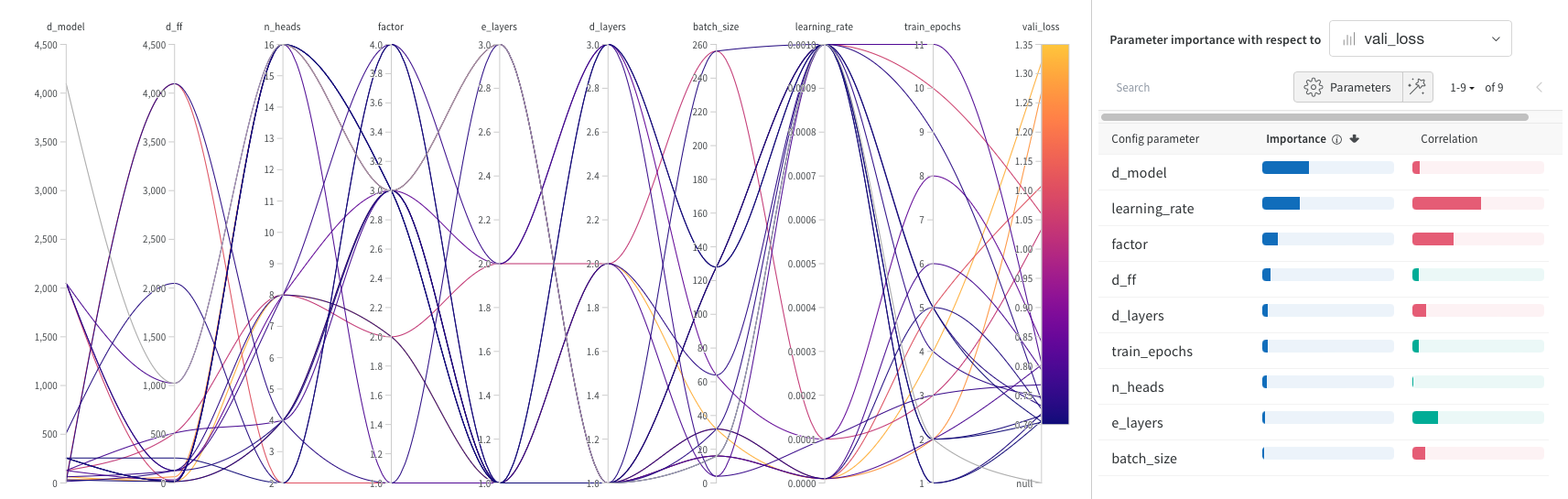}
    \caption{Crossformer}
    \label{fig:crossformer_parallel_coordinates_plot}
  \end{subfigure}
\hfill
  \begin{subfigure}[b]{0.95\linewidth}
    \includegraphics[width=\textwidth]{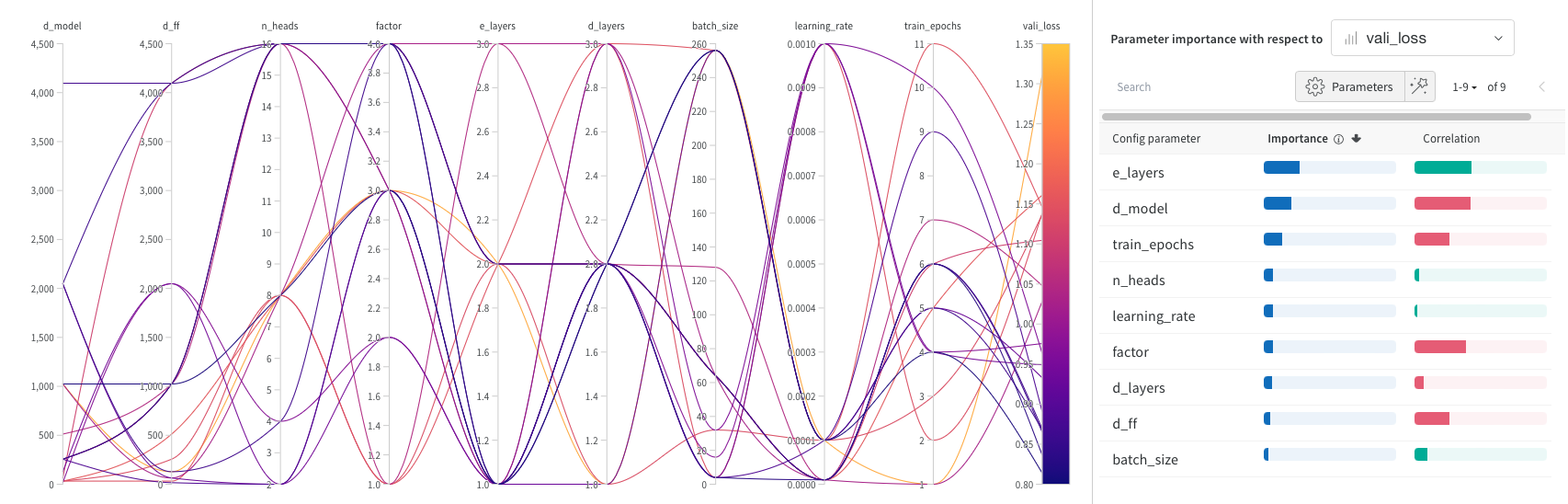}
    \caption{Non-stationary transformer}
    \label{fig:non_stationary_parallel_coordinates_plot}
  \end{subfigure}
\hfill
  \begin{subfigure}[b]{0.95\linewidth}
    \includegraphics[width=\textwidth]{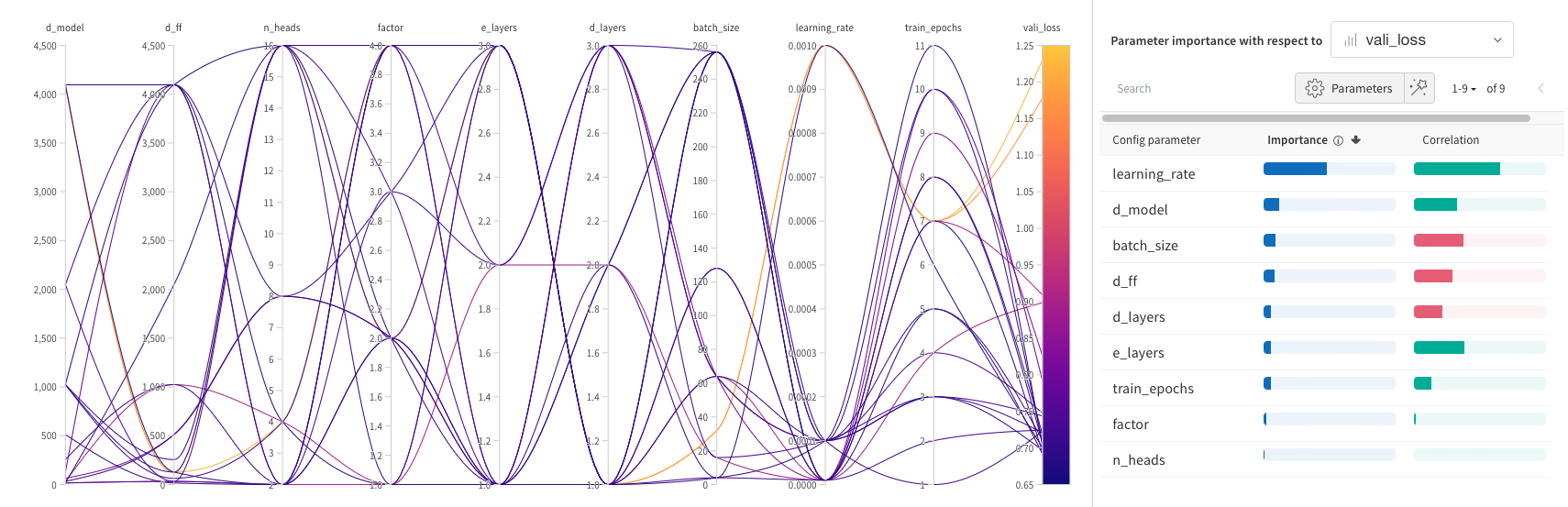}
    \caption{PatchTST}
    \label{fig:patchtst_parallel_coordinates_plot}
  \end{subfigure}
\hfill
  \begin{subfigure}[b]{0.95\linewidth}
    \includegraphics[width=\textwidth]{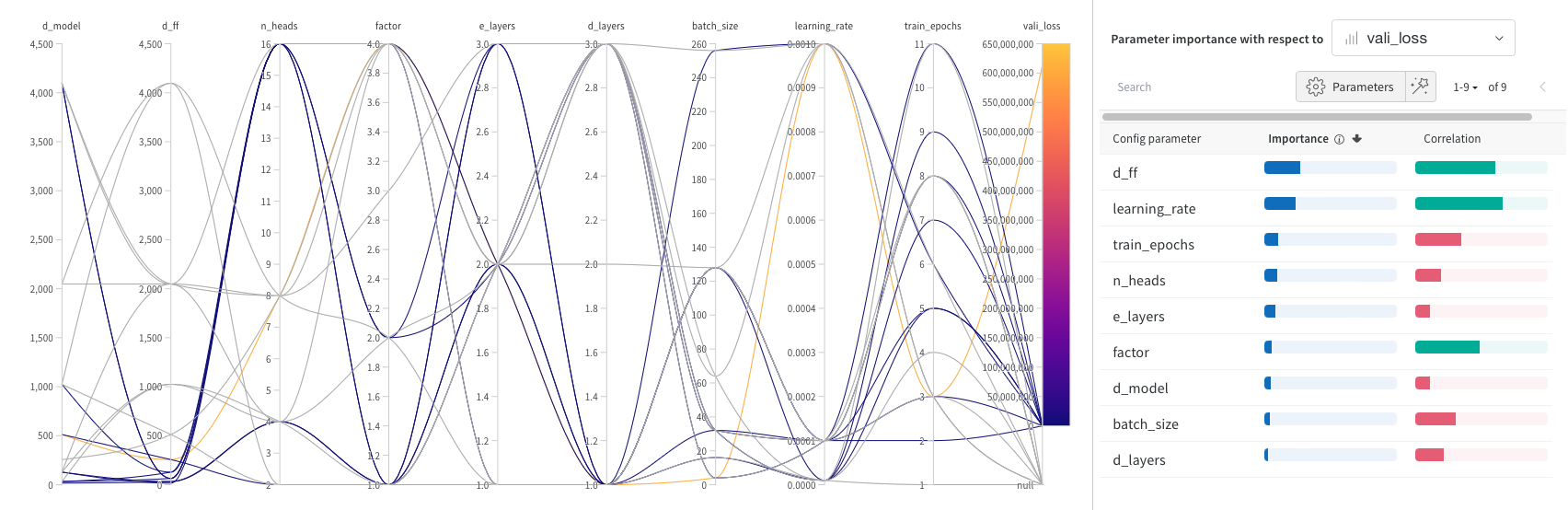}
    \caption{Mamba}
    \label{fig:mamba_parallel_coordinates_plot}
  \end{subfigure}
\hfill
  \begin{subfigure}[b]{0.95\linewidth}
    \includegraphics[width=\textwidth]{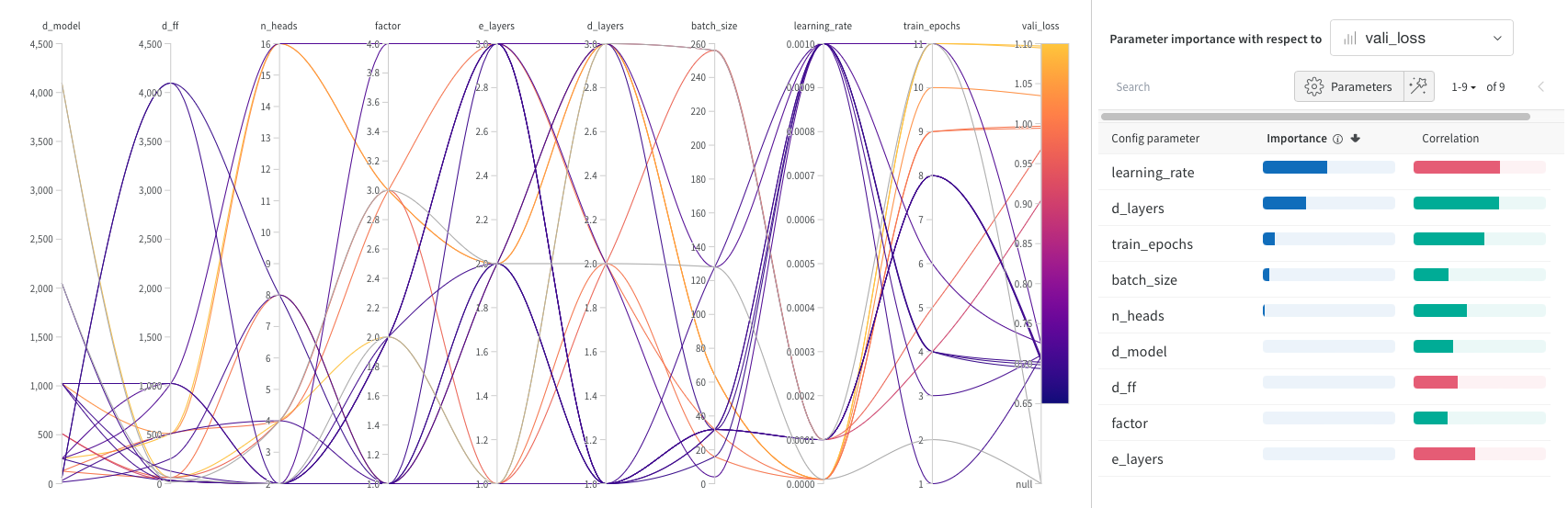}
    \caption{TimeMixer}
    \label{fig:timemixer_parallel_coordinates_plot}
  \end{subfigure}
  \caption{Parallel coordinates plot on ETTh1 dataset}
\end{figure}

\begin{figure}[h!]
\centering
  \begin{subfigure}[b]{0.95\linewidth}
    \includegraphics[width=\textwidth]{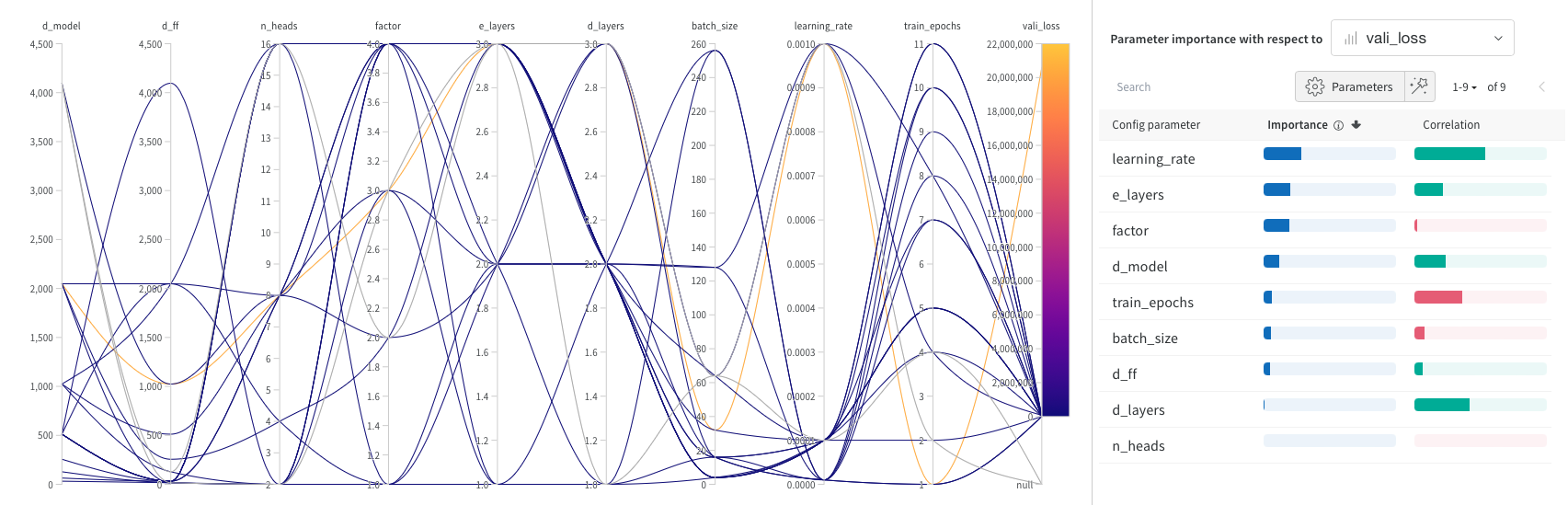}
    \caption{Autoformer}
    \label{fig:weather_autoformer_parallel_coordinates_plot}
  \end{subfigure}
\hfill
  \begin{subfigure}[b]{0.95\linewidth}
    \includegraphics[width=\textwidth]{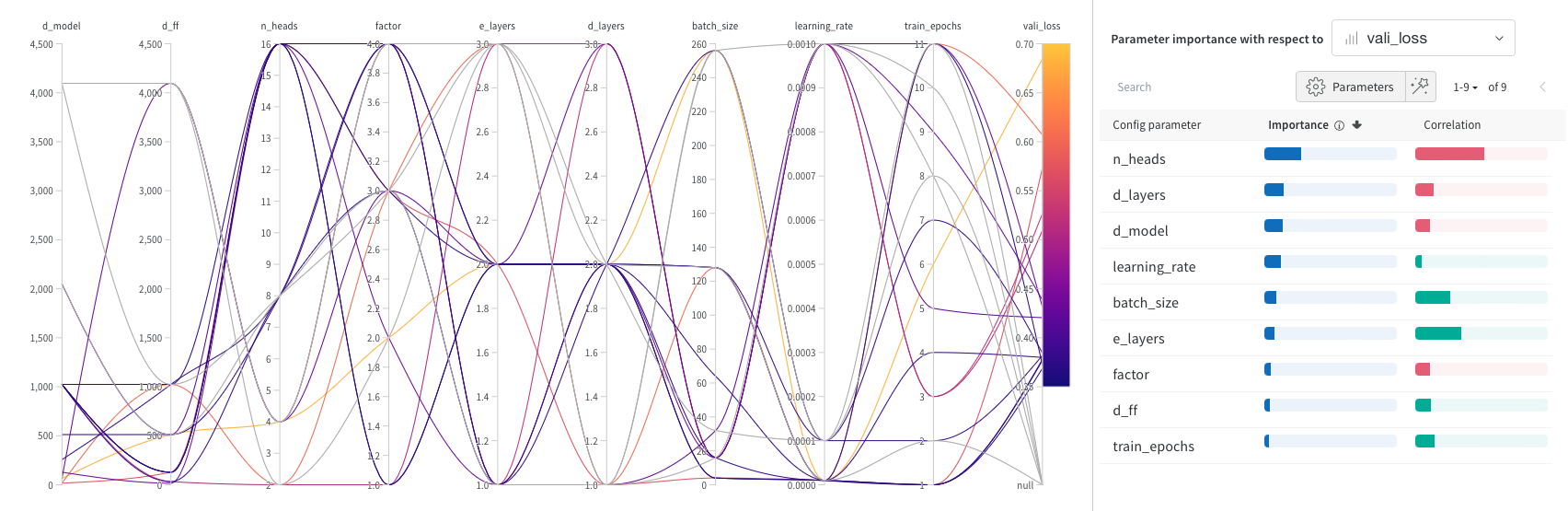}
    \caption{Crossformer}
    \label{fig:weather_crossformer_parallel_coordinates_plot}
  \end{subfigure}
\hfill
  \begin{subfigure}[b]{0.95\linewidth}
    \includegraphics[width=\textwidth]{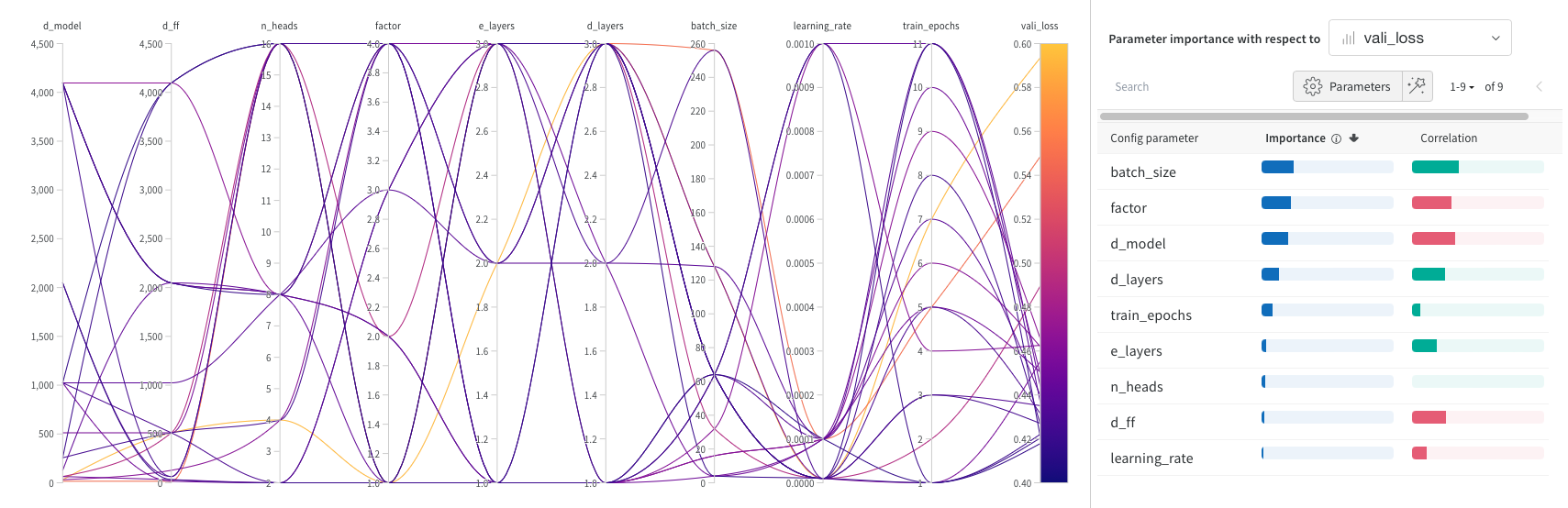}
    \caption{Non-stationary transformer}
    \label{fig:weather_non_stationary_parallel_coordinates_plot}
  \end{subfigure}
\hfill
  \begin{subfigure}[b]{0.95\linewidth}
    \includegraphics[width=\textwidth]{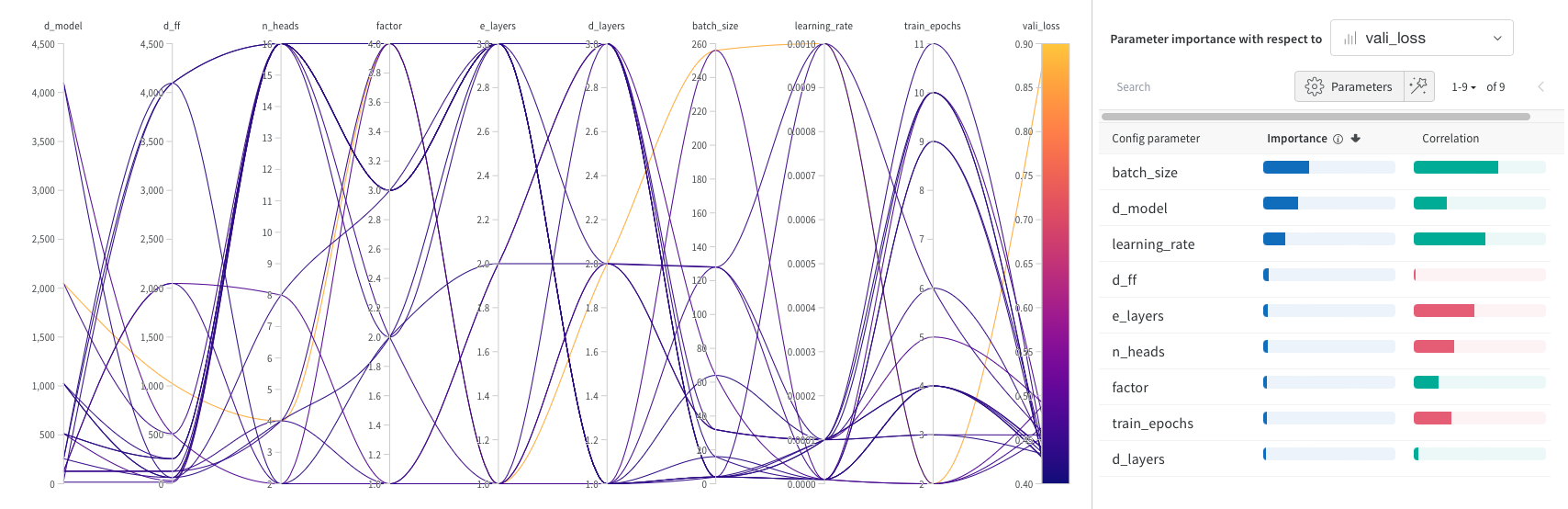}
    \caption{PatchTST}
    \label{fig:weather_patchtst_parallel_coordinates_plot}
  \end{subfigure}
\hfill
  \begin{subfigure}[b]{0.95\linewidth}
    \includegraphics[width=\textwidth]{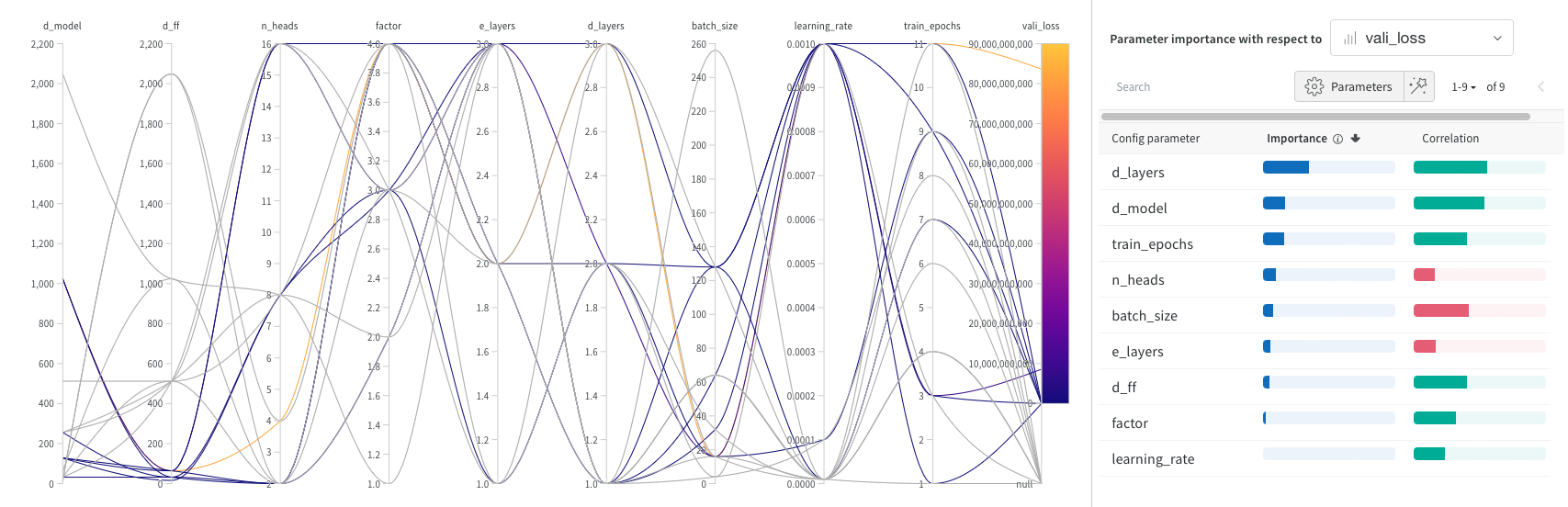}
    \caption{Mamba}
    \label{fig:weather_mamba_parallel_coordinates_plot}
  \end{subfigure}
\hfill
  \begin{subfigure}[b]{0.95\linewidth}
    \includegraphics[width=\textwidth]{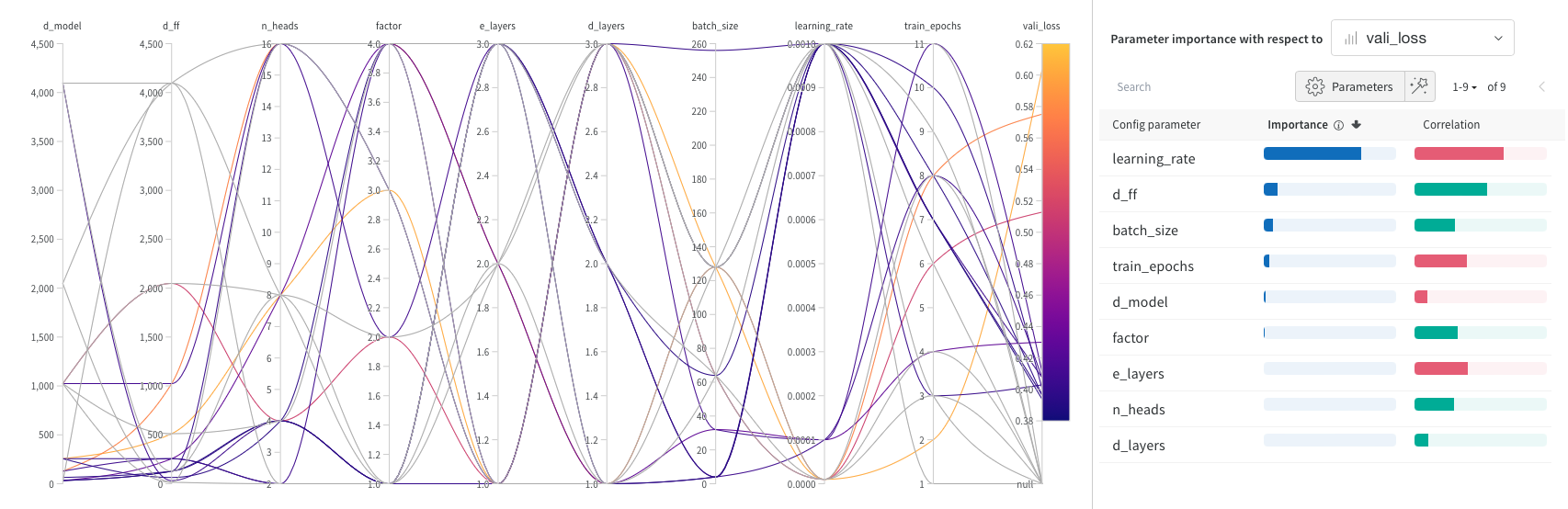}
    \caption{TimeMixer}
    \label{fig:weather_timemixer_parallel_coordinates_plot}
  \end{subfigure}
  \caption{Parallel coordinates plot on Weather dataset}
\end{figure}

ETTh1, the smallest dataset in the experiment, experienced fewer OOM errors during hyperparameter tuning. Fig.\ref{fig:autoformer_parallel_coordinates_plot} shows that for Autoformer, $batch\_size$, $train\_epoch$, and $n\_head$ are the most influential hyperparameters. Specifically, lower $batch\_size$ ($\leq 32$), higher $train\_epoch$ ($\geq 9$), and $n\_head = 4$ lead to better performance. For Crossformer (Fig.\ref{fig:crossformer_parallel_coordinates_plot}), $d\_model$, $learning\_rate$, and $factor$ are the top three hyperparameters, with smaller $d\_model (\leq 512)$, higher $learning\_rate (0.001)$, and higher $factor (\geq 3)$ resulting in better performance. In the Non-Stationary Transformer model (Fig.~\ref{fig:non_stationary_parallel_coordinates_plot}), $d\_model = 32$ and $d\_ff \leq 512$ while $n\_head = 8$ and $d\_layer = 2$ lead to higher validation loss (bad performance). PatchTST demonstrates that a smaller $learning\_rate (\leq 0.0001)$ with a higher $train\_epoch (\geq 5)$ reduce validation loss. Mamba, being very sensitive to model size, performs better when $d\_ff \leq 128$, reducing both OOM errors and validation loss. TimeMixer achieves better results with a larger $learning\_rate = 0.001$. Tab.~\ref{Tab:top3_important_parameter} summarizes the top three key parameters for each model. From the table, it is evident that $d\_model$ frequently has a significant impact on the performance of transformer-based models. In addition, different transformer-based models exhibit varying sensitivities to training-related parameters. In addition, models like Mamba and TimeMixer are predominantly influenced by training-related parameters such as the number of training epochs and learning rate.

\begin{table}[h!]
\centering
\caption{Top 3 important Parameters in each model (with respect to validation loss in ETTh1 dataset. The number means the importance rank, for example, $d\_model$ is the most influence parameter in the Crossformer model.)}
\label{Tab:top3_important_parameter}
\resizebox{\linewidth}{!}{
\begin{tabular}{c | cccccc | ccc}
\hline
\multirow{2}{*}{model} & \multicolumn{6}{c|}{Model Related}                         & \multicolumn{3}{c}{Training Related}        \\
\cline{2-10}
                       & d\_model & d\_ff & n\_head & factor & e\_layer & d\_layer & batch\_size & train\_epoch & learning\_rate \\
\hline
Autoformer             &          &       & 3       &        &          &          & 1           & 2            &                \\
Crossformer            & 1        &       &         & 3      &          &          &             &              & 2              \\
Non-s. Trans.          & 2        &       &         &        & 1        &          &             & 3            &                \\
PatchTST               & 2        &       &         &        &          &          & 3           &              & 1              \\
\hline 
Mamba                  &          & 1     &         &        &          &          &             & 3            & 2              \\
TimeMixer              &          &       &         &        &          & 2        &             & 3            & 1   \\
\hline
\end{tabular}
}
\end{table}

\subsubsection{Dataset: Weather}

The weather dataset has more variables and longer sequences compared to ETTh1, making models more prone to OOM errors, particularly in Autoformer and Mamba. To investigate further, we remove outliers from Autoformer’s results and observed that batch size is the most influential parameter (see Fig.\ref{fig:autoformer_zoomin_weather_parallel_plot}). This finding suggests that larger datasets require smaller batch sizes, which should be a focus in future hyperparameter tuning. For other models, the ranking of parameter importance from the parallel coordinate plots should guide further adjustments. For example, Crossformer’s plot (Fig.\ref{fig:weather_crossformer_parallel_coordinates_plot}) suggests that keeping $n\_head$ at a higher level $(> 8)$ during future training would improve performance.

\section{Conclusion}
\label{5_conclusion}



This paper introduces a unified hyperparameter optimization (HPO) pipeline designed for transformer-based time series forecasting (TSF) models. We benchmark several SOTA models, including Autoformer, Crossformer, Non-Stationary Transformer, PatchTST, Mamba, and TimeMixer, across multiple datasets. The tuning results highlight the significant influence of key hyperparameters such as $d\_model$, learning rate, and batch size on model performance. Our pipeline, which leverages Ray Tune and Weights \& Biases for scalable tuning and visualization, provides an efficient framework for integrating additional models. Future work will focus on expanding model coverage, exploring advanced search techniques, and addressing out-of-memory (OOM) errors through distributed hyperparameter tuning across multiple GPUs on High Performance Computing (HPC). The code and results are publicly available to facilitate ongoing research in this area.




\printbibliography
\end{document}